\documentclass[runningheads]{llncs}

\usepackage[utf8]{inputenc}
\usepackage[OT1]{fontenc}

\usepackage{graphicx}
\usepackage{amsmath,amssymb} %
\usepackage{color}
\usepackage{amsopn}

\usepackage{hyperref}
\usepackage{wrapfig}

\usepackage[capitalise]{cleveref} %
\Crefname{equation}{Equation}{Equations}
\Crefname{figure}{Figure}{Figures}
\crefformat{equation}{Equation~#2#1#3}
\crefrangeformat{equation}{Equations~#3#1#4 to~#5#2#6}
\crefmultiformat{equation}{Equations~#2#1#3}{ and~#2#1#3}{, #2#1#3}{ and~#2#1#3}

\usepackage[width=122mm,left=12mm,paperwidth=146mm,height=193mm,top=12mm,paperheight=217mm]{geometry}

\newcommand{\NEW}[1]{#1}
\newcommand{\IGNORE}[1]{}

\newcommand{\pose}{\mathbf{p}}

\newcommand{\I}{\mathcal{I}}

\begin{document}
\pagestyle{headings}
\mainmatter

\title{EgoCap: Egocentric Marker-less Motion Capture with Two Fisheye Cameras\\(Extended Abstract)\thanks{The full paper is accepted for publication at SIGGRAPH Asia 2016 \cite{Rhodin:2016egotrack}.}} %

\titlerunning{ EgoCap: Egocentric Marker-less Motion Capture (Extended Abstract)}

\authorrunning{Rhodin et al.}

\author{
		Helge Rhodin$^\text{1}$ \,\,
		Christian Richardt$^\text{1, 2, 3}$ \,\,
		Dan Casas$^\text{1}$ \,\,
		Eldar Insafutdinov$^\text{1}$ \\
		Mohammad Shafiei$^\text{1}$ \,\,
		Hans-Peter Seidel$^\text{1}$ \,\,
		Bernt Schiele$^\text{1}$ \,\,
		Christian Theobalt$^\text{1}$}

\institute{	$^\text{1}$Max Planck Institute for Informatics\quad%
	$^\text{2}$Intel Visual Computing Institute\quad%
	$^\text{3}$University of Bath}

\maketitle

\begin{abstract}
Marker-based and marker-less optical skeletal motion-cap\-ture methods use an \emph{outside-in} arrangement of cameras placed around a scene, with viewpoints converging on the center. 
They often create discomfort by possibly needed marker suits, and their recording volume is severely restricted and often constrained to indoor scenes with controlled backgrounds. 
We therefore propose a new method for real-time, marker-less and egocentric motion capture which estimates the full-body skeleton pose from a lightweight stereo pair of fisheye cameras that are attached to a helmet or virtual-reality headset.
It combines the strength of a new generative pose estimation framework for fisheye views with a ConvNet-based body-part detector trained on a new automatically annotated and augmented dataset.
Our \emph{inside-in} method captures full-body motion in general indoor and outdoor scenes, and also crowded scenes.

\end{abstract}

\section{Introduction}

Traditional optical skeletal motion-capture methods -- both marker-based and marker-less -- use several cameras typically placed around a scene in an \textit{outside-in} arrangement, with camera views approximately converging in the center of a confined recording volume.
This greatly constrains the spatial extent of motions that can be recorded; simply enlarging the recording volume by using more cameras, for instance to capture an athlete, is not scalable.
In other cases, a scene may be cluttered with objects or furniture, or other dynamic scene elements, such as people in close interaction, may obstruct a motion-captured person in the scene or create unwanted dynamics in the background.
In such cases, even state-of-the-art outside-in marker-less optical methods that succeed with just a few cameras, and are designed for outdoor scenes \cite{ElhayAJTPABST2015}, quickly fail.
This problem can partly be bypassed with motion-capture methods that use body-worn sensors.
Shiratori et al. propose to wear 16 cameras placed on body parts facing \textit{inside-out} \cite{ShiraPSSH2011}, and capture the skeletal motion through structure-from-motion relative to the environment.
This clever solution requires instrumentation, calibration and a static background, but allows free roaming and was inspirational for our egocentric approach.

\begin{figure}[t]
	\includegraphics[width=\linewidth]{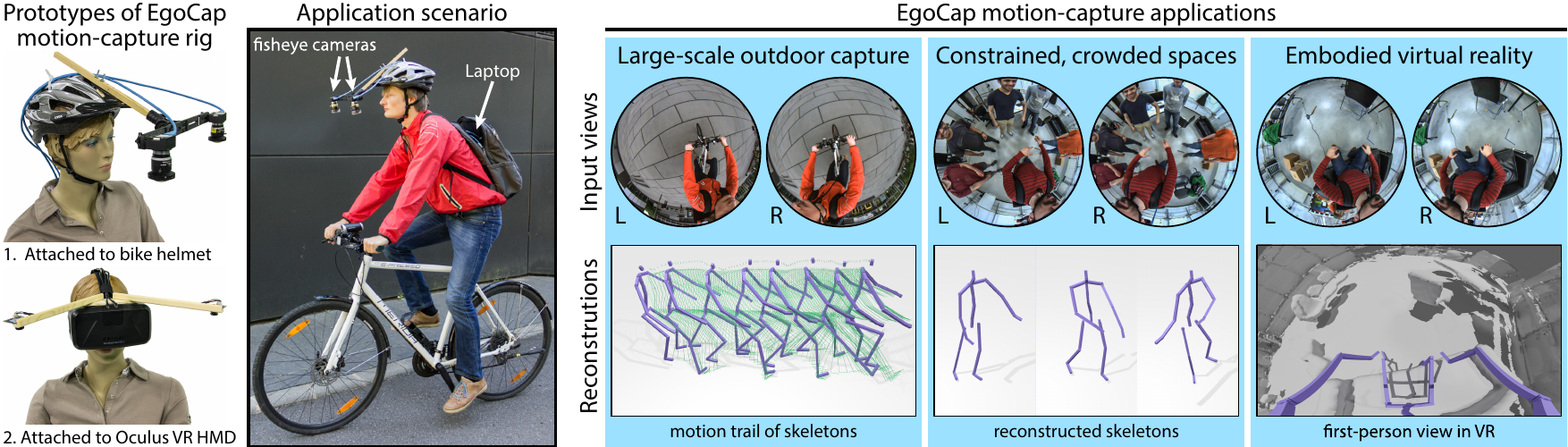}%
	\caption{
		We propose a marker-less optical motion-capture approach that only uses two head-mounted fisheye cameras (see rigs on the left).
		Our approach enables three new application scenarios:
		(1) capturing human motions in outdoor environments of virtually unlimited size,
		(2)~capturing motions in space-constrained environments, e.g. during social interactions, and
		(3) rendering the reconstruction of one's real body in virtual reality for embodied immersion.
	}
	\label{fig:teaser}
\end{figure}

We propose EgoCap: an egocentric motion-capture approach that
estimates full-body pose from a pair of optical cameras carried by
lightweight headgear (see \cref{fig:teaser}). The body-worn cameras are
oriented such that their field of view covers the user’s body entirely,
forming an arrangement that is independent of external sensors – an
 \textit{optical inside-in} method.
It reduces the setup effort, enables free roaming, and minimizes body instrumentation.
EgoCap decouples the estimation of local body pose with respect to the headgear cameras and global headgear position, which we infer by structure-from-motion on the scene. %

Our first contribution is a new egocentric inside-in sensor rig with only two head-mounted, downward-facing commodity video cameras with fisheye lenses (see \cref{fig:teaser} left).
The rig can be attached to a helmet or a head-mounted VR display, and, hence, requires less instrumentation and calibration
than other body-worn systems.
The stereo fisheye optics keep the whole body in view in all poses, despite the cameras' proximity to the body. %

\begin{wrapfigure}{R}{0.4\textwidth}%
	\vspace{-1.5\baselineskip}
	\centering%
	\includegraphics[width=0.4\textwidth]{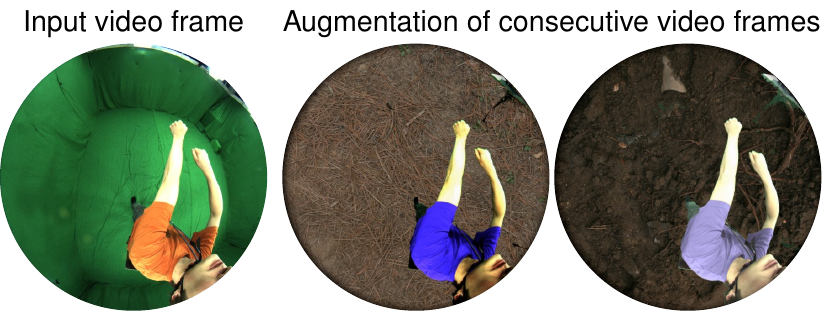}\vspace{-0.5em}
	\caption{\label{fig:augmentation}Dataset augmentation.}\vspace{-0.5cm}
\end{wrapfigure}

Our second contribution is a new marker-less motion-capture algorithm tailored to the strongly distorted egocentric fisheye views.
It combines a generative model-based skeletal pose estimation approach with evidence from a trained ConvNet-based body-part detector, and is designed to work with unsegmented frames and general backgrounds (\cref{sec:method}).

Our third contribution is a new approach for automatically creating body-part detection training datasets.
We record test subjects in front of green screen with an existing outside-in marker-less motion-capture system to get ground-truth skeletal poses, which are reprojected into the simultaneously recorded head-mounted fisheye views to get 2D body-part annotations.
We augment the training frames by replacing the green screen with random background images, and vary the appearance in terms of color and shading by intrinsic recoloring \cite{MekaZRT2016}.
With this technique, we annotate 100,000 images of egocentric videos of eight people in different clothing.
We provide the dataset for research purposes \cite{EgoCapData2016}.

\IGNORE{
\section{Related Work} %
In the past, egocentric camera placements were used for tracking or model learning of certain parts of the body, for example of the face with a helmet-mounted camera or rig \cite{JonesFYMBIBD2011,WangCF2016}, of fingers from a wrist-worn camera \cite{Kim:2012}, 
of eyes and eye gaze from cameras in a head-mounted rig \cite{Sugano:2015},
or articulated hand motion from body- or chest-worn RGB-D cameras \cite{Rogez2014,SridhMOT2015}.
Using a body-worn depth camera, \cite{YonemMOSST2015} extrapolate arm and torso poses from arm-only RGB-D footage.
\cite{JiangG2016} attempted full-body pose estimation from a chest-worn camera view by analyzing the scene, but without observing the user directly and at very restricted accuracy.
Articulated full-body motion capture with a lightweight head-mounted camera pair was not yet attempted.

A complementary research branch analyses the environment from first-person, i.e.~body-worn outward-facing cameras,
for activity recognition \cite{FathiFR2011,kitani2011fast,OhnisKKH2016,ma2016going},
for learning engagement and saliency patterns \cite{ParkJS2012,SuG2016},
and for understanding the utility of surrounding objects \cite{rhinehart2016learning}.
Articulated full-body tracking is not their goal, {but synergies of both fields appear promising}.
}

\section{Egocentric Inside-In Motion Capture}
\label{sec:method}

Our egocentric setup separates human motion capture into two subproblems: (1) local skeleton pose estimation with respect to the camera rig, and (2) global rig pose estimation relative to the environment.
Global pose is estimated with existing structure-from-motion techniques \cite{moulon2013global}.
We formulate skeletal pose estimation as an analysis-by-synthesis-style optimization problem in the pose parameters $\pose^t$, 
that maximizes the alignment of a projected 3D human body model in the left $\I_\text{left}^t$ and the right $\I_\text{right}^t$ stereo fisheye views, at each video time step $t$.
We use a hybrid alignment energy combining evidence from a generative image-formation model, as well as from a discriminative detection approach: 
\begin{align}
E(\pose^t) \!=\! E_\text{color}(\pose^t) \!+\! E_\text{detection}(\pose^t) \!+\! E_\text{pose}(\pose^t) \!+\! E_\text{smooth}(\pose^t)
\text{.}
\label{eqn:objective}
\end{align}
$E_\text{color}$ is an extension of a generative ray-casting model \cite{RhodiRRST2015} to the strongly distorted fisheye views, which provides differentiable visibility through a volumetric representation. 
$E_\text{detection}$ constrains $\pose^t$ to 2D joint detections obtained from an exiting ConvNet \cite{InsafPAAS2016}, which was fine-tuned on the previously introduced dataset.
$E_\text{pose}$ penalizes violations of anatomical joint-angle limits as well as poses deviating strongly from the rest pose, and $E_\text{smooth}$ regularizes temporal changes.

\section{Evaluation and Applications}
\label{sec:evaluation}

\paragraph{\textbf{Dataset Augmentations.}}
We first evaluate the learned body-part detectors using the percentage of correct keypoints (PCK) metric \cite{Sapp2013}
 on a validation set consisting of 1000 images of two subjects that are not part of the training set.
Background augmentation during training brings a clear improvement of 67 PCK points.
Cloth recoloring additionally improves significantly by 3 PCK points.

\paragraph{\textbf{3D Body Pose Accuracy.}}
\label{sec:comparisons}

We quantitatively evaluate the 3D body pose accuracy of our approach on ground-truth data obtained with the Captury Studio.
The average Euclidean 3D distance over all 18 joints, for which detection labels are available, is 7$\pm$1\,cm for a challenging 250-frame walking sequence with occlusions, and \NEW{7$\pm$1\,cm} on a long sequence of \NEW{750} frames of gesturing and interaction.
It meets the accuracy of outside-in approaches using 2--3 cameras \cite{ElhayAJTPABST2015}.

\paragraph{\textbf{Large-scale Motion Capture.}}
We successfully tested on a basketball sequence outdoors, which shows quick motion and close interaction, 
on an outdoor walking sequence, and on a large-scale biking sequence (\cref{fig:teaser}, third column).

\paragraph{\textbf{Constrained/Crowded Spaces.}}
We also tested EgoCap for motion capture in a crowded scene, where many spectators are interacting and occluding the tracked user from the outside (\cref{fig:teaser}, fourth column).
In such a setting, as well as in settings with many obstacles and narrow sections, outside-in motion capture, even with a dense camera system, would be difficult.

\paragraph{\textbf{Immersive VR.}}\label{sec:VR}
The EgoCap head-gear (\cref{fig:teaser}, first column) is designed to be used in virtual reality (VR) applications (\cref{fig:teaser}, last column). 
Current HMD-based systems only track the pose of the display; our approach adds motion capture of the wearer's full body, which enables a much higher level of immersion.

\section{Conclusion}

We presented EgoCap, the first approach for marker-less egocentric full-body motion capture with a head-mounted fisheye stereo rig. 
EgoCap enables motion capture of dense and crowded scenes, and reconstruction of large-scale activities that would not fit into the constrained recording volumes of outside-in motion-capture methods.
It is particularly suited for HMD-based VR applications; two cameras attached to an HMD enable full-body pose reconstruction of your own virtual body \NEW{to pave the way for immersive VR experiences and interactions}.

\section*{Acknowledgements}

This research was funded by the ERC Starting Grant project CapReal (335545).

\bibliographystyle{splncs03}
\bibliography{EgoCap_short}
\end{document}